\documentclass[a4paper,10pt]{ieeeconf}      
\IEEEoverridecommandlockouts                
\usepackage{graphicx,amsmath,amssymb,bm,color,soul}
\soulregister\cite7
\soulregister\ref7

\newcommand{\trsp}{{\scriptscriptstyle\top}}

\newcommand{\ty}[1]{{\scriptscriptstyle{#1}}}
\newcommand{\tym}[1]{{\scriptscriptstyle{\mathcal{#1}}}}

\title{\LARGE \bf
Gaussians on Riemannian Manifolds:\\Applications for Robot Learning and Adaptive Control
}

\author{Sylvain Calinon
\thanks{Sylvain Calinon is with the Idiap Research Institute, Martigny, Switzerland
{\tt\small sylvain.calinon@idiap.ch}}
\thanks{I would like to thank No\'emie Jaquier, who provided relevant suggestions for the writing and organization of the article, and who carefully proofread the manuscript.}
\thanks{This work was supported by the Swiss National Science Foundation (SNSF/DFG project TACT-HAND), and by the European Commission's Horizon 2020 Programme (MEMMO project, http://www.memmo-project.eu/, grant 780684, and DexROV project, http://www.dexrov.eu/, grant 635491).}
}

\begin{document}
\maketitle
\thispagestyle{empty}
\pagestyle{empty}

\begin{abstract}
This article presents an overview of robot learning and adaptive control applications that can benefit from a joint use of Riemannian geometry and probabilistic representations. The roles of Riemannian manifolds, geodesics and parallel transport in robotics are first discussed. Several forms of manifolds already employed in robotics are then presented, by also listing manifolds that have been underexploited but that have potentials in future robot learning applications. A varied range of techniques employing Gaussian distributions on Riemannian manifolds is then introduced, including clustering, regression, information fusion, planning and control problems. Two examples of applications are presented, involving the control of a prosthetic hand from surface electromyography (sEMG) data, and the teleoperation of a bimanual underwater robot. Further perspectives are finally discussed, with suggestions of promising research directions.
\end{abstract}

\section{INTRODUCTION}
Data encountered in robotics are characterized by simple but varied geometries, which are sometimes underexploited in robot learning and adaptive control algorithms. Such data range from joint angles in revolving articulations~\cite{Park95b}, rigid body motions~\cite{Selig05,Barfoot14}, unit quaternions to represent orientations~\cite{Zeestraten17RAL}, and symmetric positive definite matrices, which can represent sensory data processed as spatial covariances~\cite{Jaquier17IROS}, inertia~\cite{Lee18,Traversaro16}, stiffness/manipulability ellipsoids~\cite{Yoshikawa85}, as well as metrics used in the context of similarity measures. 

Moreover, many applications require these heterogeneous data to be handled altogether. Several robotics techniques employ components from the framework of Riemannian geometry. But unfortunately, this is often implemented without providing an explicit link to this framework, which can either weaken the links to other techniques or limit the potential extensions. 
This can for example be the case when computing orientation errors with a logarithmic map in the context of inverse kinematics, or when interpolating between two unit quaternions with spherical linear interpolation (SLERP). This article aims at highlighting the links between existing techniques and cataloging the missing links that could be explored in further research. These points are discussed in the context of varied robot learning and adaptive control challenges. 

Riemannian manifolds are related to a wide range of problems in machine learning and robotics. This article mainly focuses on robot learning applications based on Gaussian distributions. This includes techniques requiring uncertainty and statistical modeling to be computed on structured non-Euclidean data. The article presents an overview of existing work and further perspectives in jointly exploiting statistics and Riemannian geometry. One of the appealing use of Riemannian geometry in robotics is that it provides a principled and simple way to extend algorithms initially developed for Euclidean data to other manifolds, by efficiently taking into account prior geometric knowledge about these manifolds. 

By using Riemannian manifolds, data of various forms can be treated in a unified manner, with the advantage that existing models and algorithms initially developed for Euclidean data can be extended to a wider range of data structures. It can for example be used to revisit constrained optimization problems formulated in Euclidean space, by treating them as unconstrained problems inherently taking into account the geometry of the data. 

\begin{figure}
\centering
\includegraphics[width=\columnwidth]{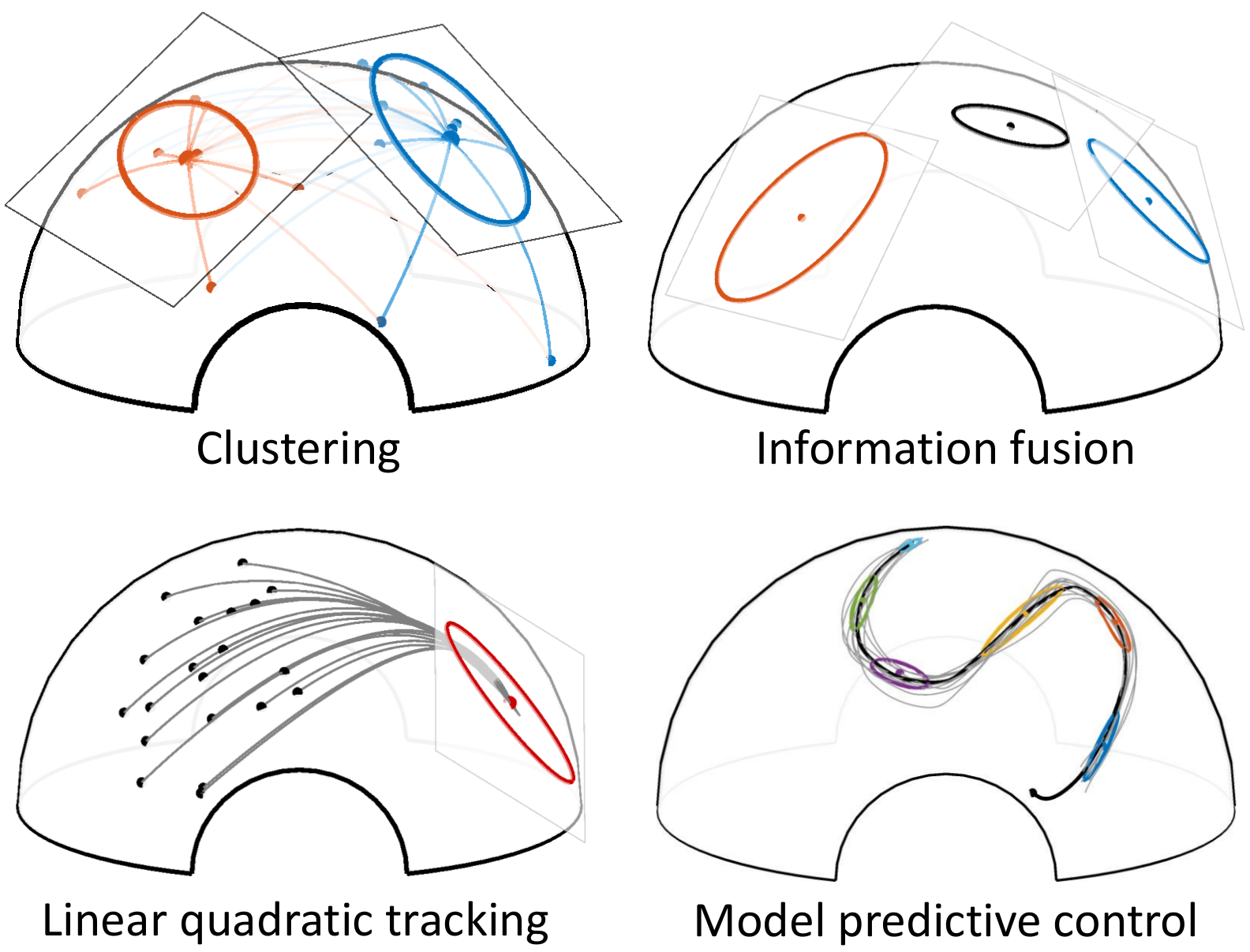}
\caption{
Problems in robotics that can leverage the proposed Gaussian-based representation on Riemannian manifolds. Such an approach can be used to extend clustering, regression, fusion, control and planning problems to non-Euclidean data (see main text for details). In these examples, Gaussians are defined with centers on the manifolds and covariances in the tangent spaces of the centers.
} 
\label{fig:problems}
\end{figure}

Figure~\ref{fig:problems} shows various common problems in robotics that can directly leverage Riemannian geometry. In the \emph{clustering} example, a set of datapoints is clustered as two distributions represented in red and blue, trained as a Gaussian mixture model, where each point is displayed in the color of the most likely cluster in which it belongs. In the \emph{information fusion} example, the intersection of two distributions (in red and blue) results in a distribution (in black), which is the equivalent of a product of Gaussians. In the \emph{tracking} example, a controller is computed by solving a linear quadratic tracking problem, with the goal of reaching a target point on the manifold within a desired precision matrix, represented here as a covariance matrix (inverse of precision matrix) in the tangent space of the target point. In the \emph{model predictive control} example, a reference path is defined as a set of Gaussians that act as viapoints to pass through (i.e., within desired covariances). This Gaussian mixture model is first learned from a set of demonstrated reference paths (in gray lines). The resulting controller computes a series of control commands anticipating the next points to reach, resulting in a path on the manifold (in black line).

The problems depicted in Fig.~\ref{fig:problems} require data to be handled in a probabilistic manner. For Euclidean data, multivariate Gaussian distributions are typically considered to encode either the (co)variations of the data or the uncertainty of the estimates. This article discusses how such approaches can be extended to other manifolds by exploiting a Riemannian extension of Gaussian distributions. A practitioner perspective is adopted, with the goal of conveying the main intuitions behind the presented algorithms, sometimes at the expense of a more rigorous treatment of each topic. Didactic source codes accompany the paper, available as part of PbDlib~\cite{pbdlib}, a collection of source codes for robot programming by demonstration (learning from demonstration), including various functionalities for statistical learning, dynamical systems, optimal control and Riemannian geometry. Two distinct versions are maintained, which can be used independently in Matlab (with full compatibility with GNU Octave) or in C++.

The paper is organized as follows. Section~\ref{sec:Riemannian} presents an overview of Riemannian geometry in robotics. Section~\ref{sec:Gaussian} presents a Gaussian-like distribution on Riemannian manifold, and shows how it can be used in mixture, regression and fusion problems. Section~\ref{sec:apps} presents examples of applications, and Section~\ref{sec:conclusion} concludes the paper.

Scalars are denoted by lower case letters $x$, vectors by boldface lower case letters $\bm{x}$, matrices by boldface uppercase letters $\bm{X}$, where $\bm{X}^\trsp$ is the transpose of $\bm{X}$.
Manifolds and tangent spaces are designated by calligraphic letters $\mathcal{M}$ and $\mathcal{T}\mathcal{M}$, respectively.

\section{RIEMANNIAN GEOMETRY IN ROBOTICS}
\label{sec:Riemannian}

\begin{figure}
\centering
\includegraphics[width=\columnwidth]{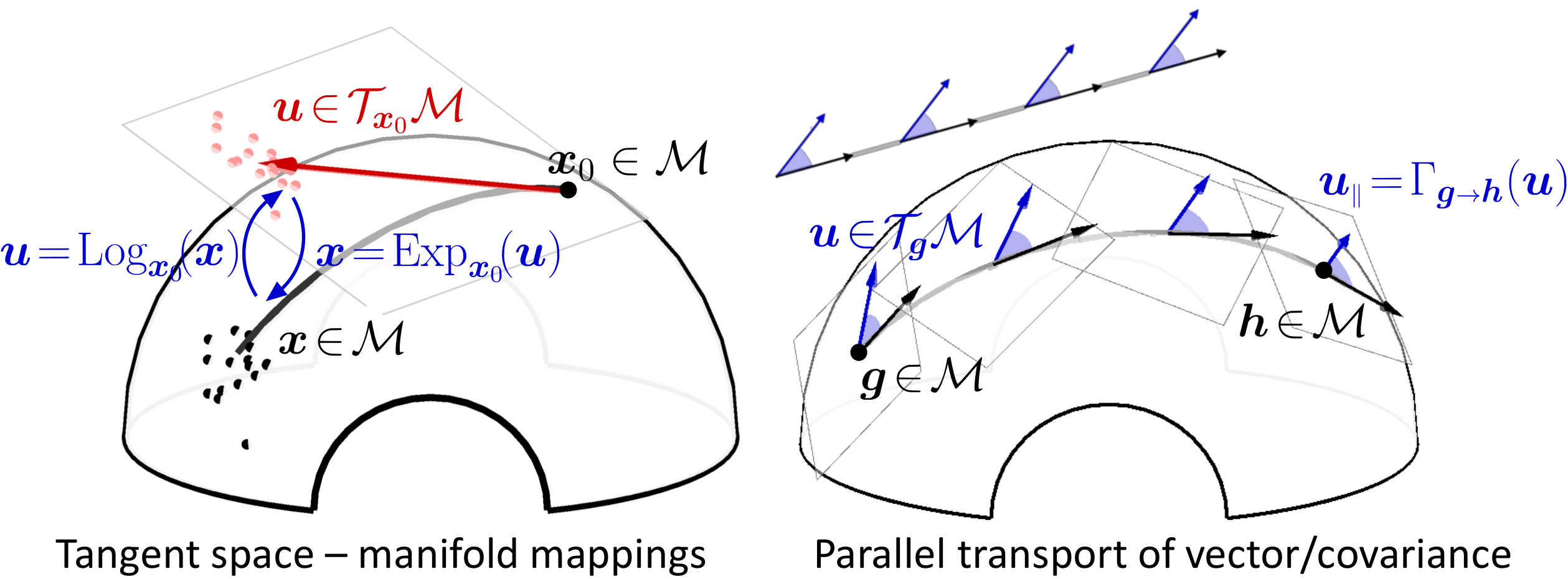}
\caption{
Applications in robotics using Riemannian manifolds rely on two well-known principles of Riemannian geometry: exponential/logarithmic mapping (\emph{left}) and parallel transport (\emph{right}), which are depicted here on a $\mathcal{S}^2$ manifold embedded in $\mathbb{R}^3$. 
\emph{Left:} Bidirectional mappings between tangent space and manifold. \emph{Right:} Parallel transport of a vector along a geodesic (see main text for details).
} 
\label{fig:functions}
\end{figure}

\subsection{Riemannian manifolds}
\label{subsec:RiemannianManifolds}
A smooth $d$-dimensional manifold $\mathcal{M}$ is a topological space that locally behaves like the Euclidean space $\mathbb{R}^d$. A \emph{Riemannian manifold} is a smooth and differentiable manifold equipped with a positive definite metric tensor. For each point $\bm{p}\!\in\!\mathcal{M}$, there exists a tangent space $\mathcal{T}_{\bm{p}} \mathcal{M}$ that locally linearizes the manifold. On a Riemannian manifold, the metric tensor induces a positive definite inner product on each tangent space $\mathcal{T}_{\bm{p}} \mathcal{M}$, which allows vector lengths and angles between vectors to be measured. The affine connection, computed from the metric, is a differential operator that provides, among other functionalities, a way to compute geodesics and to transport vectors on tangent spaces along any smooth curves on the manifold~\cite{Pennec06,Absil07}. It also fully characterizes the intrinsic curvature and torsion of the manifold. The Cartesian product of two Riemannian manifolds is also a Riemannian manifold (often called manifold bundles or manifold composites), which allows joint distributions to be constructed on any combination of Riemannian manifolds. 

Two basic notions of Riemannian geometry are crucial for robot learning and adaptive control applications, which are illustrated in Fig.~\ref{fig:functions} and described below.
\\[2mm]
\noindent\textbf{Geodesics:} 
The minimum length curves between two points on a Riemannian manifold are called geodesics. Similarly to straight lines in Euclidean space, the second derivative is zero everywhere along a geodesic. The exponential map $\text{Exp}_{\bm{x}_0}: \mathcal{T}_{\bm{x}_0} \mathcal{M}\to \mathcal{M}$ maps a point $\bm{u}$ in the tangent space of $\bm{x}_0$ to a point $\bm{x}$ on the manifold, so that $\bm{x}$ lies on the geodesic starting at $\bm{x}_0$ in the direction $\bm{u}$. The norm of $\bm{u}$ is equal to the geodesic distance between $\bm{x}_0$ and $\bm{x}$. The inverse map is called the logarithmic map $\text{Log}_{\bm{x}_0}: \mathcal{M} \to \mathcal{T}_{\bm{x}_0}\mathcal{M}$. Figure~\ref{fig:functions}-\emph{left} depicts these mapping functions.
\\[2mm]
\noindent\textbf{Parallel transport:}
Parallel transport $\Gamma_{\bm{g}\ty{\to}\bm{h}}: \mathcal{T}_{\bm{g}}\mathcal{M} \to \mathcal{T}_{\bm{h}}\mathcal{M}$ moves vectors between tangent spaces such that the inner product between two vectors in a tangent space is conserved. It employs the notion of connection, defining how to associate vectors between infinitesimally close tangent spaces. This connection allows the smooth transport of a vector from one tangent space to another by sliding it (with infinitesimal moves) along a curve.

Figure~\ref{fig:functions}-\emph{right} depicts this operation. The flat surfaces show the coordinate systems of several tangent spaces along the geodesic. The black vectors represent the directions of the geodesic in the tangent spaces. The blue vectors are transported from $\bm{g}$ to $\bm{h}$. Parallel transport allows a vector $\bm{u}$ in the tangent space of $\bm{g}$ to be transported to the tangent space of $\bm{h}$, by ensuring that the angle (i.e., inner product) between $\bm{u}$ and the direction of the geodesic (represented as black vectors) are conserved. At point $\bm{g}$, this direction is expressed as $\text{Log}_{\bm{g}}(\bm{h})$. This operation is crucial to combine information available at $\bm{g}$ with information available at $\bm{h}$, by taking into account the rotation of the coordinate systems along the geodesic (notice the rotation of the tangent spaces in Fig.~\ref{fig:functions}-\emph{right}). In Euclidean space (top-left inset), such parallel transport is simply the identity operator (a vector operation can be applied to any point without additional transformation).

By extension, a covariance matrix $\bm{\Sigma}$ can be transported with $\bm{\Sigma}_{\ty{\|}} = \sum_{i=1}^d \Gamma_{\bm{g}\ty{\to}\bm{h}}(\bm{v}_i) \; \Gamma_{\bm{g}\ty{\to}\bm{h}}^\trsp(\bm{v}_i)$, using the eigendecomposition $\bm{\Sigma} = \sum_{i=1}^d \bm{v}_i \, \bm{v}_i^\trsp$. For many manifolds in robotics, this transport operation can equivalently be expressed as a linear mapping $\bm{\Sigma}_{\ty{\|}} = \bm{A}_{\bm{g}\ty{\to}\bm{h}} \bm{\Sigma} \, \bm{A}_{\bm{g}\ty{\to}\bm{h}}^\trsp$ 
(see Supplementary Material for the computation of $\bm{A}_{\bm{g}\ty{\to}\bm{h}}$).


\subsection{Manifolds in robot applications}
\label{sec:manifolds}

\begin{figure}
\centering
\includegraphics[width=\columnwidth]{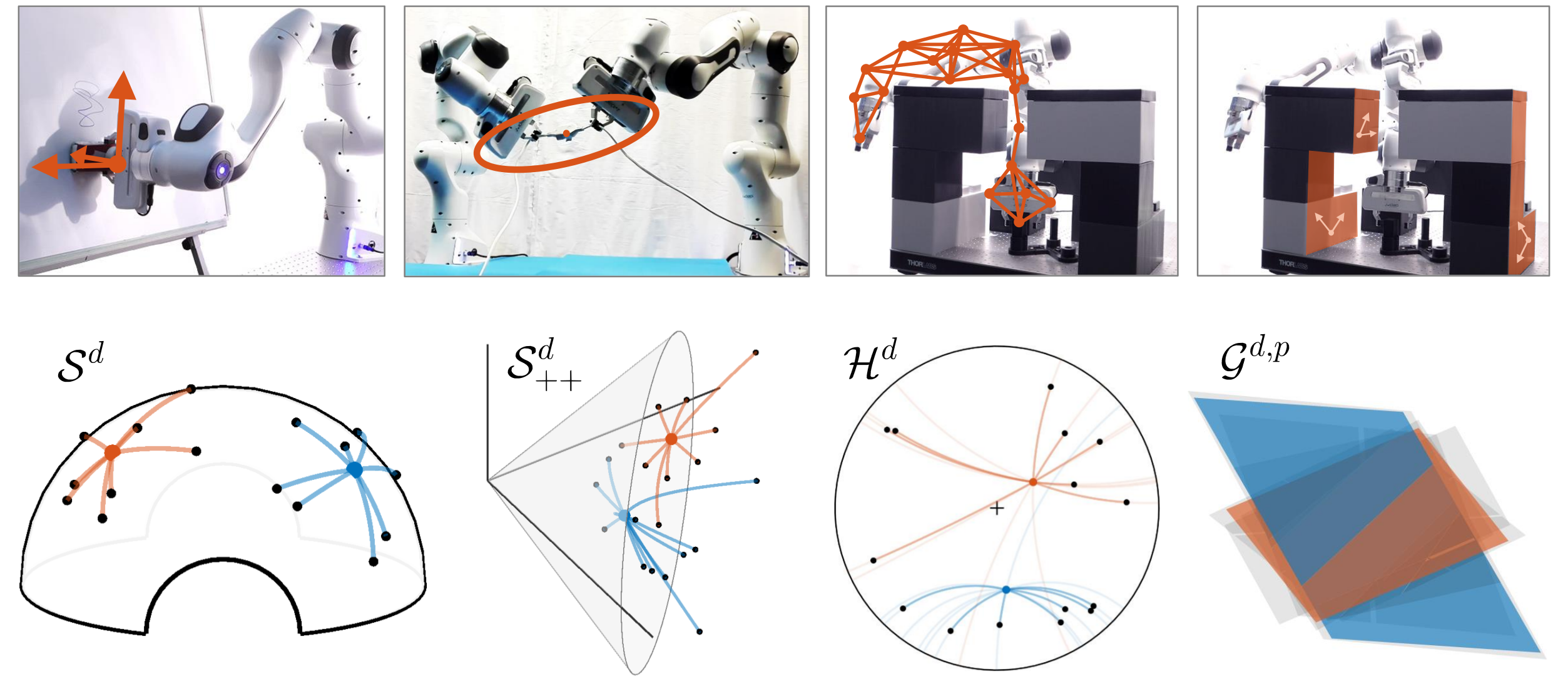}
\caption{
Structured manifolds in robotics. $\mathcal{S}^3$ can be used to represent the orientation of robot endeffectors (unit quaternions). 
$\mathcal{S}^6_{++}$ can be used to represent manipulability ellipsoids (manipulability capability in translation and rotation), corresponding to a symmetric positive definite (SPD) matrix manifold. $\mathcal{H}^d$ can be used to represent trees, graphs and roadmaps. $\mathcal{G}^{d,p}$ can be used to represent subspaces (planes, nullspaces, projection operators), see main text for details. 
%
} 
\label{fig:SSPDH}
\end{figure}

The most common manifolds in robotics are homogeneous, providing simple analytic expressions for exponential/logarithmic mapping and parallel transport. Some of the most important representations are listed below (see Supplementary Material for the corresponding mapping and transport operations). 

Figure~\ref{fig:SSPDH} shows four examples of Riemannian manifolds that can be employed in robot manipulation tasks. For these four manifolds, the bottom graphs depict $\mathcal{S}^2$, $\mathcal{S}^2_{++}$, $\mathcal{H}^2$ and $\mathcal{G}^{3,2}$, with a clustering problem in which the datapoints (black dots/planes) are segmented in two classes, each represented by a center (red and blue dots/planes).  

The geodesics depicted in Fig.~\ref{fig:SSPDH} show the specificities of each manifold. The sphere manifold $\mathcal{S}^d$ is characterized by constant positive curvature. The elements of $\mathcal{S}^d_{++}$ can be represented as the interior of a convex cone embedded in its tangent space $\text{Sym}^d$. Here, the three axes correspond to $A_{11}$, $A_{12}$ and $A_{22}$ in the SPD matrix $\left(\begin{smallmatrix} A_{11} & A_{12} \\ A_{12} & A_{22} \end{smallmatrix}\right)$. The hyperbolic manifold $\mathcal{H}^d$ is characterized by constant negative curvature. Several representations exist: $\mathcal{H}^2$ can for example be represented as the interior of a unit disk in Euclidean space, with the boundary of the disk representing infinitely remote point (Poincar\'e disk model, as depicted here). In this model, geodesic paths are arcs of circles intersecting the boundary perpendicularly. $\mathcal{G}^{d,p}$ is the Grassmann manifold of all $p$-dimensional subspaces of $\mathbb{R}^d$. 

Two Lie groups widely used in robotics are also presented, namely the special orthogonal group $\text{SO}(3)$ and the special Euclidean group $\text{SE}(3)$. Similarities and differences between Riemannian manifolds and Lie groups are discussed in the next section.
Examples of applications for the aforementioned Riemannian manifolds are presented below. 
\\[2mm]
\noindent\textbf{The sphere manifold} $\mathcal{S}^d$ can be used in robotics to encode directions/orientations. Unit quaternions $\mathcal{S}^3$ can be used to represent endeffector (tool tip) orientations~\cite{Zeestraten17RAL}. $\mathcal{S}^2$ can be used to represent unit directional vector perpendicular to surfaces (e.g., for contact planning). Articulatory joints can be represented on the torus $\mathcal{S}^1 \times \mathcal{S}^1 \times \ldots \times \mathcal{S}^1$~\cite{Arimoto09,Biess11}. The Kendall shape space used to encode 3D skeletal motion capture data also relies on unit spheres~\cite{Hosni18}.
\\[2mm]
\noindent\textbf{The special orthogonal group} $\text{SO}(d)$ is the group of rotations around the origin in a $d$-dimensional space. $\text{SO}(2)$ and $\text{SO}(3)$ are widely used in robotics. For example, in~\cite{Forster17}, the manifold structure of the rotation group is exploited for preintegration and uncertainty propagation in $\text{SO}(3)$. This is exploited for state estimation in visual-inertial odometry with mobile robots. In~\cite{Forbes17}, Kalman filtering adapted to data in $\text{SO}(3)$ is used for estimating the attitude of robots that can rotate in space. The optimization problem in~\cite{Brossette18} uses sequential quadratic programming (SQP) working directly on the manifold $\text{SO}(3) \times \mathbb{R}^3$.
\\[2mm]
\noindent\textbf{The special Euclidean group} $\text{SE}(3)$ is the group of rigid body transformations. A rigid body transformation is composed of a rotation and a translation. The geometry of $\text{SE}(3)$ can be used to describe the kinematics and the Jacobian of robots~\cite{Selig05}. Therefore, it is widely used to describe robot motion and pose estimation~\cite{Zefran96}. For example, in~\cite{Barfoot14}, exponential maps are exploited to associate uncertainty with $\text{SE}(3)$ datapoints in robot pose estimation problems. 
\\[2mm]
\noindent\textbf{The manifold of symmetric positive definite (SPD) matrices} $\mathcal{S}_{++}^d$ can be employed in various ways in robotics. For example, human-robot collaboration applications require the use of various sensors. These sensory data can be preprocessed with sliding windows to analyze at each time step the evolution of the signals within a short time window (e.g., to analyze data flows). Often, such analysis takes the form of spatial covariances, which are SPD matrices~\cite{Jaquier17IROS}. In robot control, tracking gains can be defined in the form of SPD matrices. The use of tracking gains as full SPD matrices instead of scalars has the advantage of allowing the controller to take into account the coordination of different control variables (e.g., motor commands). For articulatory joints, these coordinations often relate to characteristic synergies in human movements. Manipulability ellipsoids are representations used to analyze and control the robot dexterity as a function of the articulatory joints configuration. This descriptor can be designed according to different task requirements, such as tracking a desired position or applying a specific force~\cite{Yoshikawa85,Jaquier18}. Manipulator inertia matrices also belong to $\mathcal{S}_{++}^d$ and can for example be exploited in humanlike trajectory planning~\cite{Biess11}. SPD matrices are also used in problems related to metric interpolation/extrapolation and metric learning~\cite{Hauberg12}. 
In CHOMP (covariant Hamiltonian motion planning)~\cite{Zucker13}, a precision matrix (metric tensor) is used to prefer perturbations resulting in small accelerations in the overall trajectory. 
In~\cite{Cheng18}, Riemannian manifold policies are employed to generate natural obstacle-avoiding reaching motion through traveling along geodesics of curved spaces defined by the presence of obstacles.  
\\[2mm]
\noindent\textbf{Hyperbolic manifolds} $\mathcal{H}^d$ are the analogues of spheres with constant negative curvature instead of constant positive curvature. They are currently underexploited in robotics, despite their interesting potential in a wide range of representations, including dynamical systems, Toeplitz/Hankel matrices or autoregressive models~\cite{Chevallier15}. Hyperbolic geometry could notably be used to encode and visualize heterogeneous topology data, including graphs and trees structures, such as rapidly exploring random trees (RRT)~\cite{LaValle01}, designed to efficiently search nonconvex, high-dimensional spaces in motion planning by randomly building a space-filling tree. The interesting property of hyperbolic manifolds is that the circumference of a circle grows exponentially with its radius, which means that exponentially more space is available with increasing distance. It provides a convenient representation for hierarchies, which tend to expand exponentially with depth. 
\\[2mm]
\noindent\textbf{Grassmannian} $\mathcal{G}^{d,p}$ is the manifold of all $p$-dimensional subspaces of $\mathbb{R}^d$. It can for example be used to extract and cluster planar surfaces in the robot's 3D environment. This manifold is largely underrepresented in robotics, despite such structure can be used in various approaches such as system identification~\cite{Usevich14}, spatiotemporal modeling of human gestures~\cite{Slama15}, or the encoding of nullspaces and projection operators in a probabilistic way. 
\\[2mm]

\noindent\textbf{Manifolds with nonconstant curvature} are also employed in robotics, such as spaces endowed with the Fisher information metric~\cite{Wilson14,Amari16} or kinetic energy metric~\cite{Park95b,Zefran96,Arimoto09,Biess11}. 
As described in Section~\ref{subsec:RiemannianManifolds}, the curvature of a Riemannian manifold depends on the selected metric tensor. Consequently, a varying metric will result in a varying curvature. Many problems in robotics can be formulated with a smoothly varying matrix $\bm{M}$ (Riemannian metric) that measures the distance between two points $\bm{x}_1$ and $\bm{x}_2$ as a quadratic error term $(\bm{x}_1\!-\!\bm{x}_2)^\trsp\bm{M}(\bm{x}_1\!-\!\bm{x}_2)$. In this context, the Riemannian formulation has the advantage of being coordinate independent (i.e., geodesic paths are invariant to the choice of local coordinates)~\cite{Park95b,Arimoto09,Biess11}. In robot dynamics problems, this is typically useful to take into account the inertia in the robot motion~\cite{Park95b}. 
In policy learning problems, if the conditional density of the action given the state is Gaussian, the natural policy gradient is given by the Fisher information matrix~\cite{Amari16}, which can for example be used in deep reinforcement learning~\cite{Rajeswaran18}. 

It is also relevant for deep generative models such as variational autoencoders (VAEs) and generative adversarial networks (GANs), as it provides a geometric interpretation of these models. For example, VAEs learn nonlinear data distributions through a set of latent variables and a nonlinear generator function that maps latent points into the input space. The nonlinearity of the generator implies that the latent space gives a distorted view of the input space. The latent space not only provides a low-dimensional representation of the data manifold: it can also reveal its underlying geometrical structure~\cite{Arvanitidis18}.

In neural networks such as VAEs, by using activation functions that are $C^2$ differentiable, a symmetric positive definite matrix $\bm{M}=\bm{J}^\trsp\bm{J}$ can be used as a smoothly changing inner product structure, acting as a local Mahalanobis distance measure, where $\bm{J}$ is the Jacobian characterizing the decoder function. The method yields a distance measure that can for example be used to replace linear interpolations in the latent space by geodesics. In~\cite{Arvanitidis18}, Arvanitidis \emph{et al.} show that in the latent space learned by a VAE, distances and interpolants can significantly be improved under this metric, which in turn improves probability distributions, sampling algorithms and clustering in the latent space.

A downside of manifolds with nonconstant curvature is that the geodesic optimization problem most often corresponds to a nonconvex problem (system of ordinary differential equations) that can be computationally heavy to solve. Several research directions can be explored to address this issue. In~\cite{Chen19}, the problem is circumvented by spanning the latent space with a discrete finite graph ($k$-d tree data structure with edge weights based on Riemannian distances), used in conjunction with a classic $A^*$ search algorithm. Recent approaches in discrete differential geometry also address similar problems by extending continuous Riemannian manifolds to discrete formulations with fast computation~\cite{Sharp19}. Currently, these developments principally target computer graphics applications, but they have great potentials in robotics. It could for example provide an approach to link discrete robot planning problems such as probabilistic roadmaps (PRMs) to their continuous counterparts in Riemannian geometry.

\subsection{Riemannian geometry and Lie theory}
\label{sec:Lie}
A Lie group is a smooth and differentiable manifold that possesses a group structure, therefore satisfying the group axioms. There are strong links between Riemannian geometry and Lie theory. In particular, some Lie groups, such as $\text{SO}(3)$, can be endowed with a bi-invariant Riemannian metric, which give them the structure of a Riemannian manifold. In robotics, Lie theory is mainly exploited for applications involving $\text{SO}(3)$ and $\text{SE}(3)$ groups.

In the literature, distinctive vocabulary and notation are often employed, which hinder some of the links between the applications exploiting Riemannian geometry and Lie theory.
Among these differences, the \emph{Lie algebra} is the tangent space at the origin of the manifold, acting as a global reference. $\bm{u}^\wedge$ (\emph{hat}) and $\bm{u}^\vee$ (\emph{vee}) are used to transform elements from the Lie algebra (which can have nontrivial structures such as complex numbers or skew-symmetric matrices) to vectors in $\mathbb{R}^d$, which are easier to manipulate. They are the operations corresponding to the exponential and logarithm maps in Riemannian geometry. In Lie theory, $\oplus$ and $\ominus$ are operators used to facilitate compositions with exponential/logarithmic mapping operations.

For further reading, an excellent introduction to Lie theory for robot applications can be found in~\cite{Sola19}.

\section{GAUSSIAN DISTRIBUTIONS ON RIEMANNIAN MANIFOLDS}
\label{sec:Gaussian}
Several approaches have been proposed to extend Gaussian distributions in Euclidean space to Riemannian manifolds~\cite{Said17}. Here, we focus on a simple approach that consists of estimating the mean of the Gaussian as a centroid on the manifold (also called Karcher/Fr\'echet mean), and representing the dispersion of the data as a covariance expressed in the tangent space of the mean~\cite{Pennec06,SimoSerra16,Zeestraten17RAL}. 
Distortions arise when points are too far apart from the mean, but this distortion is negligible in most robotics applications. In particular, this effect is strongly attenuated when a mixture of Gaussians is considered, as each Gaussian will be employed to model a limited region of the manifold. In the general case of a manifold $\mathcal{M}$, such a model is a distribution maximizing the entropy in the tangent space. It is defined as
\begin{equation*}
	\mathcal{N}_{\mathcal{M}}(\bm{x}|\bm{\mu},\bm{\Sigma}) = 
	{\Big((2\pi)^{d} | \bm{\Sigma} |\Big)}^{-\frac{1}{2}} \; 
	e^{-\frac{1}{2} \text{Log}_{\bm{\mu}}\!(\bm{x}) \, \bm{\Sigma}^{-1} \, \text{Log}_{\bm{\mu}}\!(\bm{x})},
\end{equation*}
where $\bm{x}\!\in\!\mathcal{M}$ is a point of the manifold, $\bm{\mu}\!\in\!\mathcal{M}$ is the mean of the distribution (origin of the tangent space), and $\bm{\Sigma}\in\mathcal{T}_{\bm{\mu}}\mathcal{M}$ is the covariance defined in this tangent space.

For a set of $N$ datapoints, this geometric mean corresponds to the minimization 
\begin{equation*}
	\min_{\bm{\mu}} \sum_{n=1}^N {\text{Log}_{\bm{\mu}}\!(\bm{x}_n)}^\trsp \; \text{Log}_{\bm{\mu}}\!(\bm{x}_n),
\end{equation*}
which can be solved by a simple and fast Gauss-Newton iterative algorithm. The algorithm starts from an initial estimate on the manifold and an associated tangent space. The datapoints $\{\bm{x}_n\}_{n=1}^N$ are projected in this tangent space to compute a direction vector, which provides an updated estimate of the mean. This process is repeated by iterating
\begin{equation*}
	\bm{u} = \frac{1}{N} \sum_{n=1}^N \text{Log}_{\bm{\mu}}\!(\bm{x}_n),
	\qquad
	\bm{\mu} \leftarrow \text{Exp}_{\bm{\mu}}\!(\bm{u}),
\end{equation*}
until convergence. In practice, such an algorithm converges very fast in only a couple of iterations (typically less than 10 for the accuracy required by the applications presented here). After convergence, a covariance is computed in the tangent space as $\bm{\Sigma}=\frac{1}{N} \sum_{n=1}^N \text{Log}_{\bm{\mu}}\!(\bm{x}_n) \, \text{Log}_{\bm{\mu}}^\trsp\!(\bm{x}_n)$, see Fig.~\ref{fig:problems}. This distribution can for example be used in a control problem to represent a reference to track with an associated required precision (e.g., learned from a set of demonstrations). Such a learning and control problem results in the linear quadratic tracking (LQT) solution depicted in Fig.~\ref{fig:problems} and described in details in~\cite{Zeestraten17IROS}.

Importantly, this geometric mean can be directly extended to weighted distances, which will be exploited in the next sections for mixture modeling, fusion (product of Gaussians), regression (Gaussian conditioning) and planning problems.

\subsection{Gaussian mixture model}
\label{sec:GMM}

\begin{figure}
\centering
\includegraphics[width=\columnwidth]{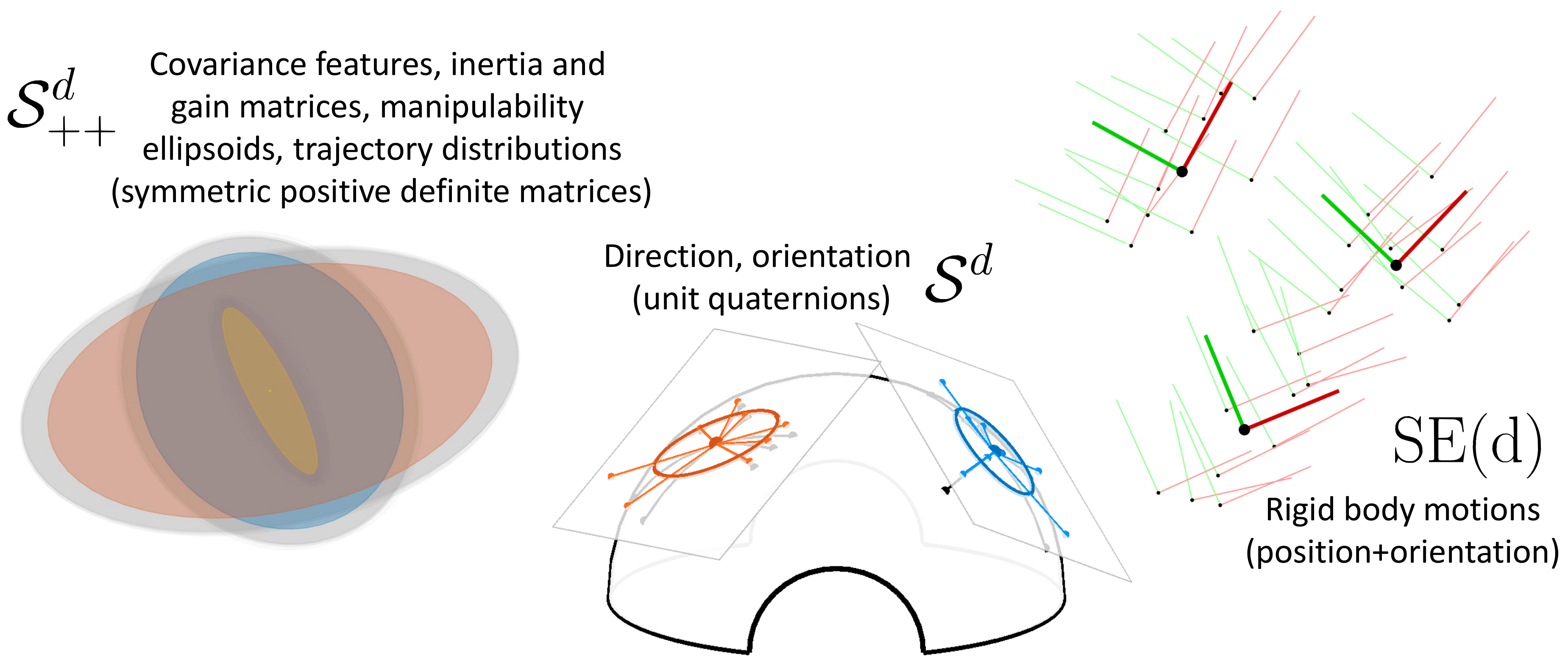}
\caption{
Clustering on various manifolds with Gaussian mixture models. 
} 
\label{fig:clustering}
\end{figure}

\begin{figure}
\centering
\includegraphics[width=\columnwidth]{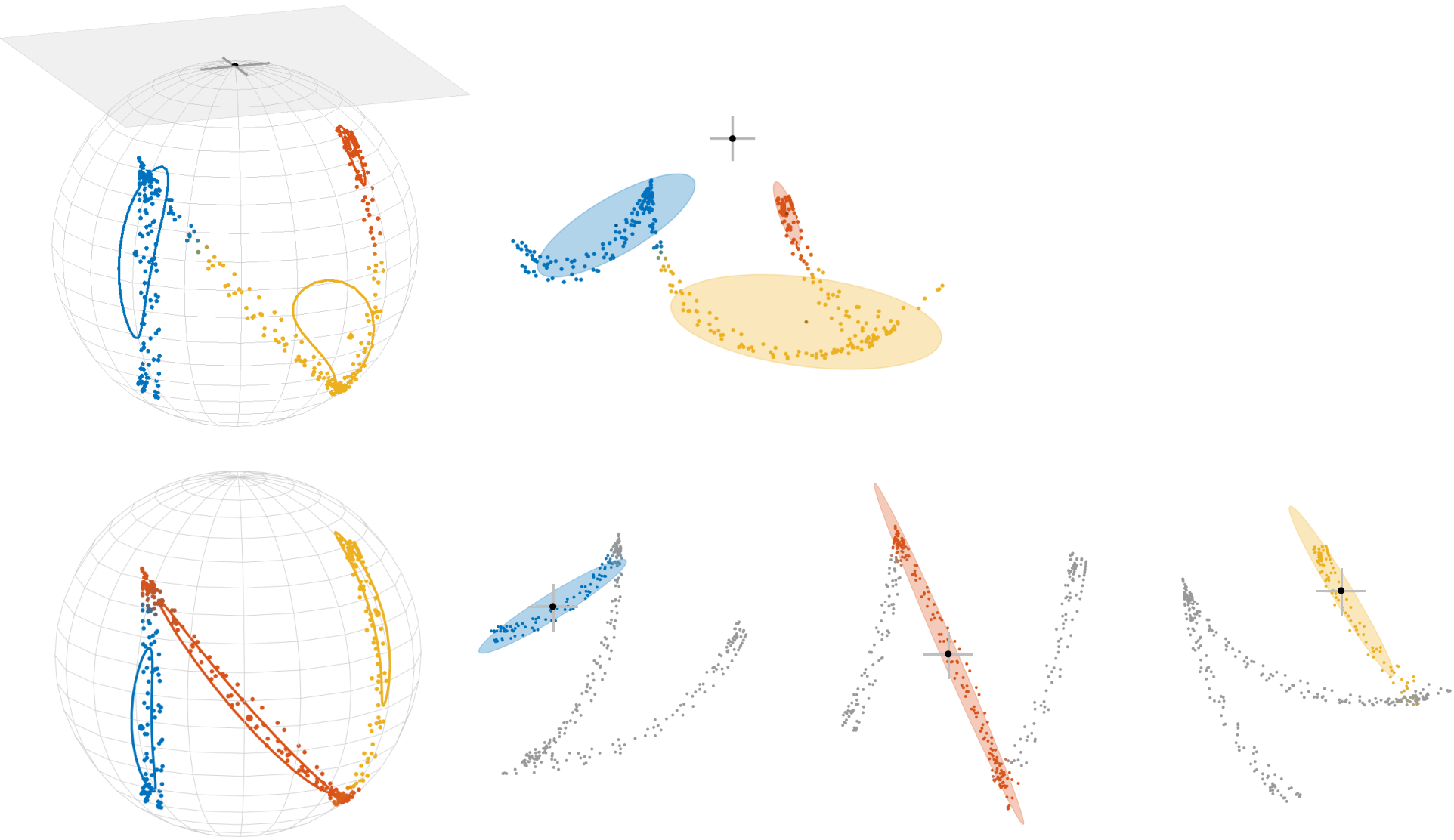}
\caption{
Gaussian mixture model (GMM) trained with EM algorithm on $\mathcal{S}^2$ manifold. \emph{Top row:} GMM encoding in a single tangent space (at the origin). \emph{Bottom row:} Proposed GMM, where the covariances are computed in the tangent spaces of the means. On the left figures, the contours of the covariances are projected on the manifold. The right figures show the projections of the data into the tangent spaces considered in the computation of the GMM. We can see that representing the local dispersion of the data as covariances in the tangent spaces of the means (bottom row) results in a much better fit than representing the GMM in a single tangent space (top row). 
} 
\label{fig:GMM}
\end{figure}

Gaussian mixture model (GMM) is a ubiquitous representation in robotics, including clustering and modeling of distributions as a superposition of Gaussians, see Fig.~\ref{fig:clustering}. Similarly to a GMM in the Euclidean space, a GMM on a manifold $\mathcal{M}$ is defined by $p(\bm{x}) = \sum_{k=1}^{K} \pi_k \, \mathcal{N}_{\mathcal{M}}(\bm{x}|\bm{\mu}_k,\bm{\Sigma}_k)$, with $K$ the number of components and $\pi_k$ the mixing coefficients (priors) such that $\sum_k \pi_k\!=\!1$ and $\pi_k\geq0$, $\forall\,k\in\{1,\ldots,K\}$. The parameters of this GMM can be estimated by Expectation-Maximization (EM)~\cite{Ghahramani94}, where the Gauss-Newton procedure presented above is performed in the M-step. 

Figure \ref{fig:GMM}-\emph{top} shows that a GMM computed in a single tangent space (here, at the origin of the manifold) introduces distortions resulting in a poor modeling of the data. Figure \ref{fig:GMM}-\emph{bottom} shows that the proposed representation limits the distortions by encoding the local spread of the data in covariance matrices expressed in different tangent spaces (i.e., at the centers of the Gaussians). 

An example of application with links to robotics is~\cite{SimoSerra16}, where human poses are modeled using a GMM on $\mathcal{S}^d$. 
Matlab examples {\scriptsize\texttt{demo\_Riemannian\_Sd\_GMM*.m}} can be found in~\cite{pbdlib}.

\subsection{Gaussian conditioning}
\label{sec:conditioning}
As detailed in~\cite{Zeestraten17RAL}, we consider input and output data jointly encoded as a multivariate Gaussian $\mathcal{N}_{\mathcal{M}}(\bm{\mu},\bm{\Sigma})$ partitioned with symbols $^\tym{I}$ and $^\tym{O}$ (input and output).
Given an input datapoint $\bm{x}^\tym{I}$, the conditional distribution $\bm{x}^\tym{O}|\bm{x}^\tym{I} \sim \mathcal{N}_{\mathcal{M}}(\bm{\hat{\mu}}^\tym{O},\bm{\hat{\Sigma}}^\tym{O})$ can be locally evaluated by iterating
\begin{equation*}
	\bm{u} = \text{Log}_{\bm{\hat{\mu}}^\tym{O}}\!(\bm{\mu}^\tym{O}) - 
	\bm{\Sigma}_\ty{\|}^\tym{OI} \; {\bm{\Sigma}_\ty{\|}^\tym{I}}^{-1} \; \text{Log}_{\bm{x}^\tym{I}}\!(\bm{\mu}^\tym{I}),
	\;
	\bm{\hat{\mu}}^\tym{O} \leftarrow \text{Exp}_{\bm{\hat{\mu}}^\tym{O}}\!(\bm{u}),
\end{equation*}
with $\bm{\Sigma}_\ty{\|}$ a covariance matrix transported from ${[{\bm{\mu}^\tym{I}}^\trsp, \bm{\mu}{^\tym{O}}^\trsp]}^\trsp$ to ${[{\bm{x}^\tym{I}}^\trsp, \bm{\hat{\mu}}{^\tym{O}}^\trsp]}^\trsp$ (see Section~\ref{sec:Riemannian} for the description of parallel transport). After convergence, the covariance is computed in the tangent space as $\bm{\hat{\Sigma}}^\tym{O} = \bm{\Sigma}_\ty{\|}^\tym{O} - \bm{\Sigma}_\ty{\|}^\tym{OI} \; {\bm{\Sigma}_\ty{\|}^\tym{I}}^{-1} \bm{\Sigma}_\ty{\|}^\tym{IO}$.
Matlab examples {\scriptsize\texttt{demo\_Riemannian\_Sd\_GMR*.m}} can be found in~\cite{pbdlib}.

\subsection{Fusion with products of Gaussians}
\label{sec:fusion}
As shown in~\cite{Zeestraten17IROS,Zeestraten18RAL}, the product of $K$ Gaussians on a Riemannian manifold can be locally evaluated by iterating
\begin{equation*}
	\bm{u} = {\Bigg(\sum_{k=1}^K\bm{\Sigma}_{\ty{\|}k}^{-1}\Bigg)}^{-1} 
	\sum_{k=1}^K \bm{\Sigma}_{\ty{\|}k}^{-1} \, \text{Log}_{\bm{\mu}}\!(\bm{\mu}_k),
	\qquad
	\bm{\mu} \leftarrow \text{Exp}_{\bm{\mu}}\!(\bm{u}),
\end{equation*}
with covariance matrix $\bm{\Sigma}_{\ty{\|}k}$ transported from $\bm{\mu}_k$ to $\bm{\mu}$ (see Section~\ref{sec:Riemannian} for the description of parallel transport).
After convergence, the covariance is computed in the tangent space as $\bm{\Sigma} = {\Big(\sum_{k=1}^K \bm{\Sigma}_{\ty{\|}k}^{-1}\Big)}^{-1}$.

An example of product of Gaussians on $\mathcal{S}^2$ is depicted in the top-right inset of Fig.~\ref{fig:problems}.
A Matlab example {\scriptsize\texttt{demo\_Riemannian\_Sd\_GaussProd01.m}} can be found in~\cite{pbdlib}.

\subsection{Model predictive control}
\label{sec:MPC}

\begin{figure}
\centering
\includegraphics[width=\columnwidth]{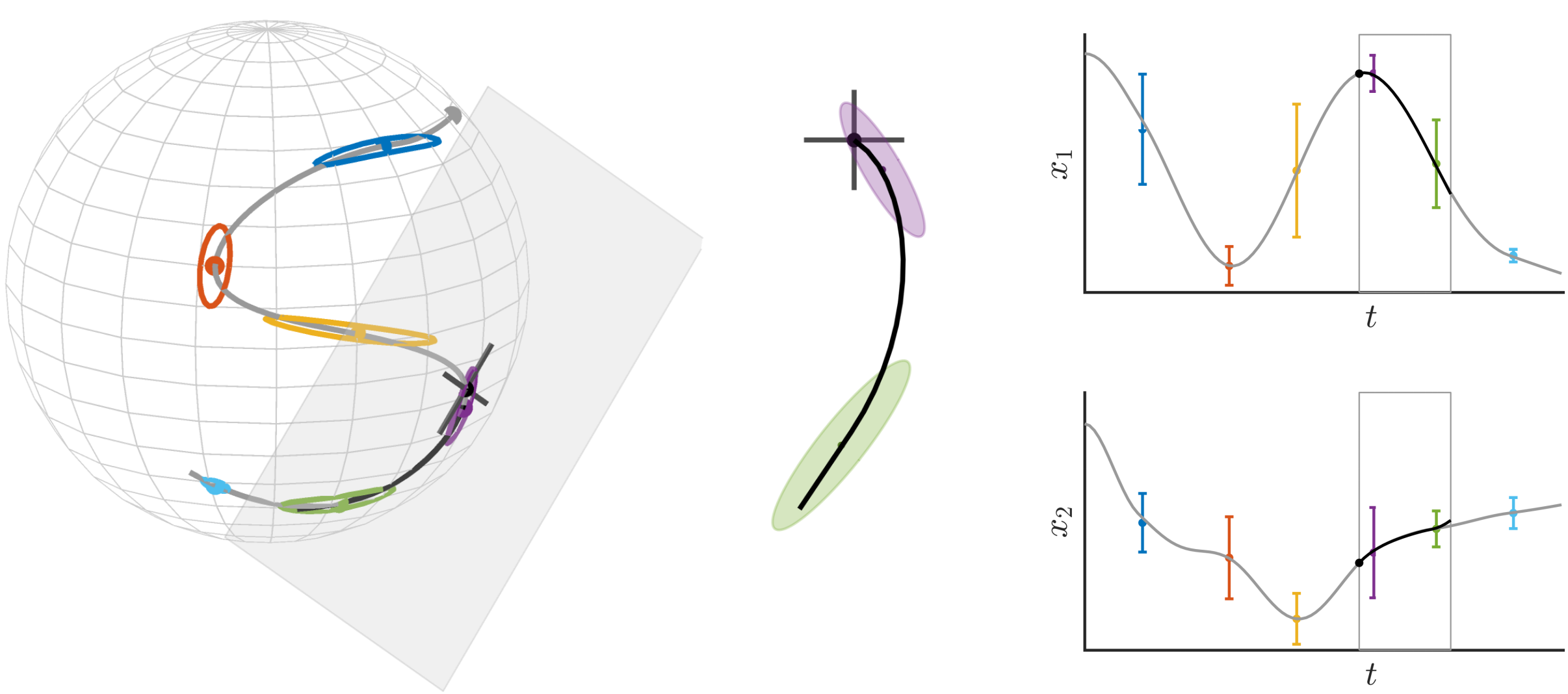}
\caption{
Model predictive control (MPC) on a $\mathcal{S}^2$ manifold, with a set of viapoints defined by a Gaussian mixture model. \emph{Left:} Final movement generated by MPC (in gray), superposed with the partial movement (in black) for the given time horizon (1/5 of total duration), predicted at 3/5 of the trajectory. \emph{Center:} Visualization in the tangent space of $\bm{x}$, where only two Gaussians appear within the current time horizon. \emph{Right:} Timeline plot showing the evolution of the first two variables of the state space, where the time horizon is depicted as a gray box. The reference to track is represented as a set of colored viapoints with desired error margins (represented as standard deviations). 
} 
\label{fig:MPC}
\end{figure}

Model predictive control (MPC) is widely employed in robotics as an adaptive control strategy with anticipation capability. It consists of estimating a series of control commands $\bm{u}$ across a moving time window of size $T\!-\!1$. The problem is described here as a linear quadratic tracking (LQT) problem with velocity commands $\bm{u}_t\in\mathbb{R}^d$ and an evolution of the state $\bm{x}_t\in\mathbb{R}^d$ described by a linear system $\bm{x}_{t+1}=\bm{A}_t\bm{x}_t+\bm{B}_t\bm{u}_t$, but the approach can be generalized to other controllers. The resulting controller is
\begin{align}
	\bm{\hat{u}} &= \arg\min_{\bm{u}} {\big\|\bm{x}-\bm{\mu}\big\|}_{\bm{Q}}^2 
	+ {\big\|\bm{u}\big\|}_{\bm{R}}^2
	\nonumber\\
	&= {\big({\bm{S}_{\bm{u}}}^\trsp \bm{Q} \bm{S}_{\bm{u}} + \bm{R}\big)}^{-1}
	{\bm{S}_{\bm{u}}}^\trsp \bm{Q} 
	\big(\bm{\mu} - \bm{S}_{\bm{x}} \bm{x}_1 \big),
	\label{eq:MPC}
\end{align}
with $\bm{x}\!=\!{\begin{bmatrix}\bm{x}_1^\trsp, \bm{x}_2^\trsp, \ldots, \bm{x}_T^\trsp \end{bmatrix}}^\trsp\!\in\!\mathbb{R}^{dT}$ the evolution of the state variable, $\bm{u}\!=\!{\begin{bmatrix}\bm{u}_1^\trsp, \bm{u}_2^\trsp, \ldots, \bm{u}_{T-1}^\trsp \end{bmatrix}}^\trsp\!\in\!\mathbb{R}^{d(T-1)}$ the evolution of the control variable, and $d$ the dimension of the state space. $\bm{\mu}\!=\!{\begin{bmatrix}\bm{\mu}_1^\trsp, \bm{\mu}_2^\trsp, \ldots, \bm{\mu}_T^\trsp \end{bmatrix}}^\trsp\!\in\!\mathbb{R}^{dT}$ represents the evolution of the reference to track, $\bm{Q}\!=\!\mathrm{blockdiag}(\bm{Q}_1,\bm{Q}_2,\ldots,\bm{Q}_T)\in\mathbb{R}^{dT\times dT}$ represents the evolution of the required tracking precision, and $\bm{R}\!=\!\mathrm{blockdiag}(\bm{R}_{1},\bm{R}_{2},\ldots,\bm{R}_{T-1})\in\mathbb{R}^{d(T-1)\times d(T-1)}$ represents the evolution of the cost on the control inputs. In~\eqref{eq:MPC}, $\bm{S}_{\bm{u}}$ and $\bm{S}_{\bm{x}}$ are transfer matrices, see Supplementary Material for details of computation. 
This formulation corresponds to a basic form of MPC in Euclidean space, by considering quadratic objective functions, and linear systems with velocity commands and position states. We showed in~\cite{Calinon16JIST} that the reference signal to be tracked can be represented by a GMM to form a stepwise trajectory. 

Equation~\eqref{eq:MPC} is typically used to compute a series of control commands across a time window, which are reevaluated at each iteration. Thus, only the first (few) commands are used in practice. In the above formulation, the first time step of this moving time window corresponds to the current time step in which the problem is solved (see Fig.~\ref{fig:MPC}-\emph{right} for an illustration of this moving time window and the computed control commands within this time window).

Such an MPC/LQT problem can be extended to Riemannian manifolds by exploiting the tangent space of the state $\bm{x}_1$ (the point that will introduce the least distortions). By extension of \eqref{eq:MPC}, we can solve at each iteration 
\begin{align}
	\bm{\hat{u}} &= \arg\min_{\bm{u}} {\big\|\text{Log}_{\bm{x}_1}\!({\bm{x}})-\text{Log}_{\bm{x}_1}\!({\bm{\mu}})\big\|}_{\bm{Q}_\ty{\|}}^2 
	+ {\big\|\bm{u}\big\|}_{\bm{R}}^2
	\nonumber\\
	&= {\big({\bm{S}_{\bm{u}}}^\trsp \bm{Q}_\ty{\|} \bm{S}_{\bm{u}} + \bm{R}\big)}^{-1}
	{\bm{S}_{\bm{u}}}^\trsp \bm{Q}_\ty{\|} \,
	\text{Log}_{\bm{x}_1}\!({\bm{\mu}}), 
	\label{eq:MPC2}
\end{align}
where the vector $\bm{\hat{u}}$ is composed of $T\!-\!1$ commands expressed in the tangent space of $\bm{x}_1$. $\text{Log}_{\bm{x}_1}\!({\bm{x}})$ and $\text{Log}_{\bm{x}_1}\!({\bm{\mu}})$ are vectors respectively composed of $T$ elements $\text{Log}_{\bm{x}_1}\!({\bm{x}_t})$ and $\text{Log}_{\bm{x}_1}\!({\bm{\mu}_{s_t}})$, with $\{s_t\}_{t=1}^T$ the sequence of Gaussian identifiers used to build the stepwise reference trajectory from the GMM. $\bm{Q}_\ty{\|}$ is a matrix composed of the block-diagonal elements 
$\bm{Q}_{\ty{\|}t} = \sum_{i=1}^{d} \Gamma_{\bm{\mu}_{s_t}\!\ty{\to}\bm{x}}(\bm{v}_i) \; \Gamma_{\bm{\mu}_{s_t}\!\ty{\to}\bm{x}}^\trsp(\bm{v}_i)$, using the eigendecomposition $\bm{Q}_{s_t} = \sum_{i=1}^{d} \bm{v}_i \, \bm{v}_i^\trsp$. This transport operation can equivalently be expressed as a linear mapping $\bm{Q}_{\ty{\|}t} = \bm{M}_{\ty{\|}} \, \bm{Q}_{s_t} \, \bm{M}_{\ty{\|}}^\trsp$. In the above formulation, $\bm{R}$ is assumed to be isotropic and, thus, does not need to be transported. 

The first velocity command in~\eqref{eq:MPC2} (denoted by $\bm{\hat{u}}_{1:d}$) is then used to update the state with
\begin{align}
	\bm{x}_1 \leftarrow \text{Log}_{\bm{x}_1}\!(\bm{B}_1\bm{\hat{u}}_{1:d}),
	\label{eq:MPC2up}
\end{align}
where $\bm{B}_1$ belongs to the linear system $\bm{x}_2=\bm{A}_1\bm{x}_1+\bm{B}_1\bm{u}_1$ at the first time step of the time window.

Figure~\ref{fig:MPC} shows an example on $\mathcal{S}^2$, where the computations in~\eqref{eq:MPC2} and~\eqref{eq:MPC2up} are repeated at each time step to reproduce a movement (with the reference to track encoded as a GMM). Extensions to more elaborated forms of MPC follow a similar principle. 
A Matlab example {\scriptsize\texttt{demo\_Riemannian\_Sd\_MPC01.m}} can be found in~\cite{pbdlib}.

\section{EXAMPLES OF APPLICATIONS}
\label{sec:apps}

The operations presented in the previous sections (mixture modeling, conditioning and fusion) can be combined in different ways, which is showcased here by two examples of applications.

\subsection{Control of prosthetic hands with Gaussian mixture regression}
\label{sec:TACTHAND}

\begin{figure}
\centering
\includegraphics[width=\columnwidth]{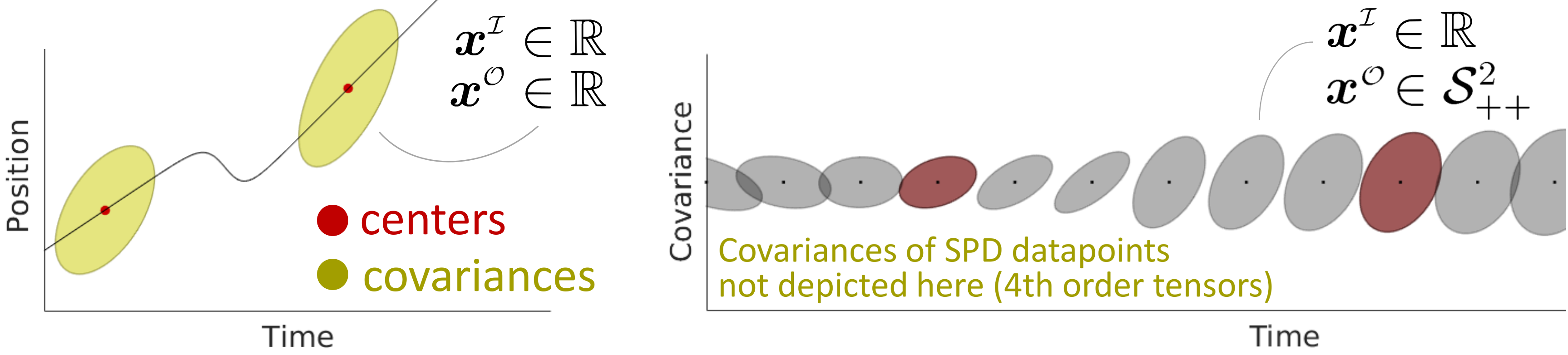}
\caption{
Gaussian mixture regression (GMR) on SPD manifold. \emph{Left:} Classical use of GMR to encode trajectories with time as input and position as output (both in the Euclidean space). \emph{Right:} Extension to Riemannian manifolds with outputs on the SPD manifold. This nonlinear regression approach provides a conditional estimate of the output expressed in the form of matrix-variate Gaussians.
} 
\label{fig:GMR}
\end{figure}

\begin{figure}
\centering
\includegraphics[width=\columnwidth]{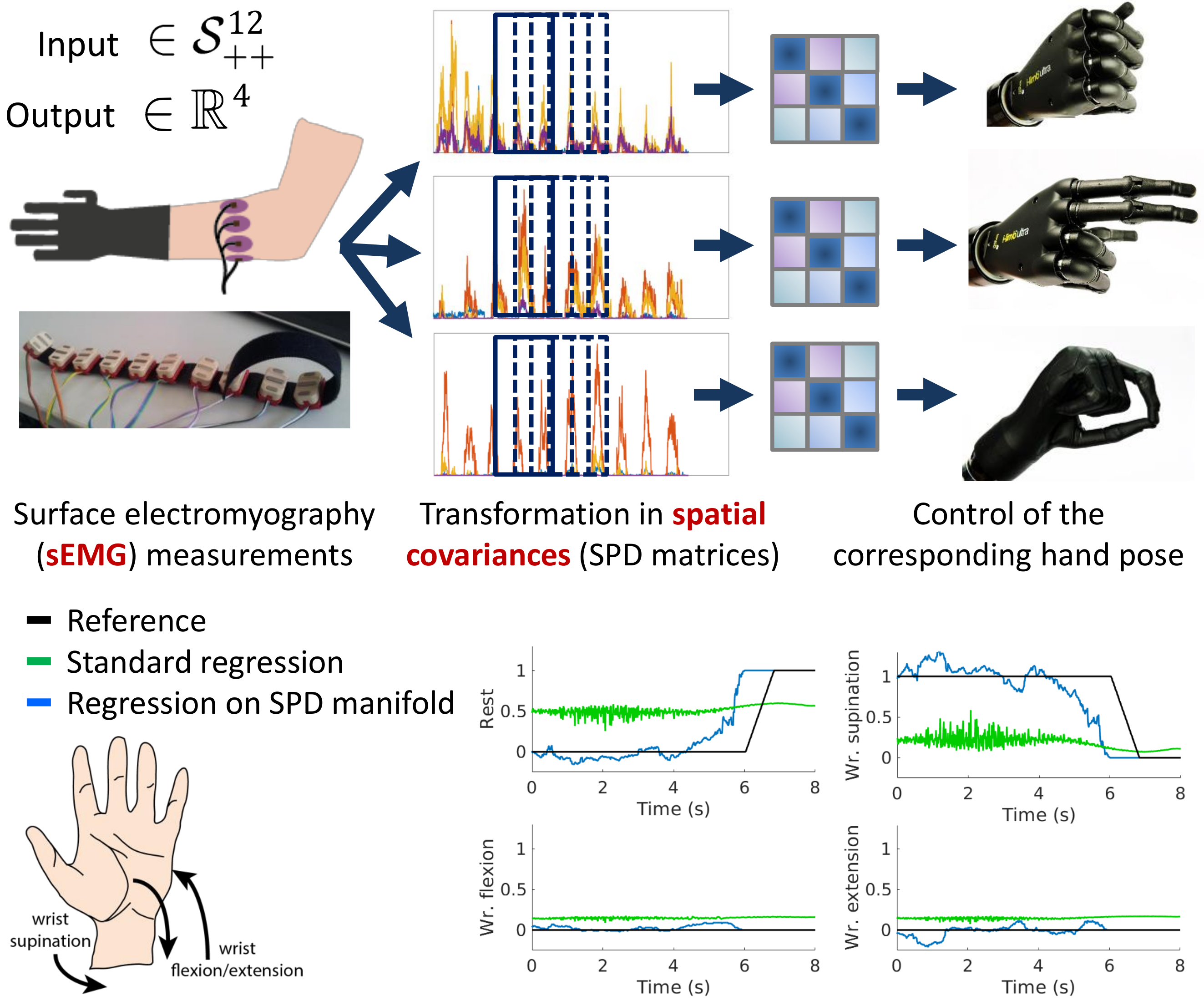}
\caption{
GMR for the control of prosthetic hands within the TACT-HAND project. SPD signals are used as input, in the form of spatial covariances computed from sEMG sensors on the forearm of the participants. Activation signals corresponding to different hand poses are used as outputs. In this experiment (see~\cite{Jaquier17IROS} for details), taking the geometry of the data into account in GMR (bottom graphs, in blue) results in better discrimination than treating the data as if they were in a Euclidean space (bottom graphs, in green).} 
\label{fig:TACTHAND}
\end{figure}

The Gaussian conditioning approach presented in Section~\ref{sec:conditioning} can be extended to the Gaussian mixture model approach presented in Section~\ref{sec:GMM}.
The resulting approach is called Gaussian mixture regression (GMR), a simple nonlinear regression technique that does not model the regression function directly, but instead first models the joint probability density of input-output data in the form of a GMM~\cite{Ghahramani94,Calinon16JIST}.  
GMR provides a fast regression approach in which multivariate output distributions can be computed in an online manner, with a computation time independent of the number of datapoints used to train the model, by exploiting the learned joint density model. In GMR, both inputs and outputs can be multivariate, and after learning, any subset of input-output dimensions can be selected for regression. This is exploited in robotics to handle different sources of missing data, where expectations on the remaining dimensions can be computed as a multivariate distribution. These properties make GMR an attractive tool for robotics, which can be used in a wide range of problems and that can be combined fluently with other techniques~\cite{Calinon16JIST}.

Both~\cite{SimoSerra16} and~\cite{Kim17} present methods for regression from a mixture of Gaussians on Riemannian manifolds, but they only partially exploit the manifold structure in Gaussian conditioning. In~\cite{SimoSerra16}, each distribution is located on its own tangent space, with the covariances encoded separately, resulting in a block-diagonal structure in the joint distribution. In~\cite{Kim17}, a GMM is reformulated to handle the space of rotation in $\mathbb{R}^3$ by using logarithm and exponential transformations on unit quaternions, with these operations formulated in a single tangent space (at the origin) instead of applying the transformations locally (see Fig.~\ref{fig:GMM}). The link to Riemannian manifolds is also not discussed.


Here, it is proposed to extend GMR to input and/or output data on SPD manifolds, see Fig.~\ref{fig:GMR}. As the covariance of SPD datapoints is a 4th-order tensor, a method is proposed in~\cite{Jaquier17IROS} for parallel transport of high-order covariances on SPD manifolds, by exploiting the supersymmetry properties of these 4th-order tensors. As an example of application, GMR on SPD manifold is applied to predict wrist movement from spatial covariances computed from surface electromyography (sEMG) data. In this application, the input data of GMR are spatial covariances that belong to the SPD manifold. Compared to the Euclidean GMR, the GMR on SPD manifold improved the detection of wrist movement for most of the participants and proved to be efficient to detect transitions between movements, see Fig.~\ref{fig:TACTHAND} and Table~\ref{Tab:TACTHAND} for a summary of the results. This shows the importance and benefits of considering the underlying manifold structure of the data in this application. The details of this experiment can be found in~\cite{Jaquier17IROS}.

\begin{table}[t]
	\caption{Comparison of the root mean square error (RMSE) obtained by GMR on the SPD manifold and the standard Euclidean GMR for wrist motion estimation from sEMG (see~\cite{Jaquier17IROS} for details). The results are presented for three participants.}
	\label{Tab:TACTHAND}
	\vspace{-4mm}
	\begin{center}
		\begin{tabular}{c|c|c|c|c|}
			& Rest & Wr. supination & Wr. extension & Wr. flexion\\
			\hline
			$\mathcal{S}^D_{++}$ & $0.29\pm0.00$ & $0.18\pm0.00$ & $0.25\pm0.00$ & $0.27\pm0.00$ \\
			$\mathbb{R}$ & $0.47\pm0.00$ & $0.31\pm0.00$ & $0.33\pm0.00$ & $0.33\pm0.00$ \\
			\hline
			$\mathcal{S}^D_{++}$ & $0.32\pm0.02$ & $0.29\pm0.14$ & $0.36\pm0.07$ & $0.43\pm0.13$ \\
			$\mathbb{R}$ & $0.46\pm0.00$ & $0.34\pm0.00$ & $0.35\pm0.00$ & $0.35\pm0.00$ \\
			\hline
			$\mathcal{S}^D_{++}$ & $0.36\pm0.02$ & $0.22\pm0.00$ & $0.31\pm0.00$ & $0.29\pm0.00$ \\
			$\mathbb{R}$ & $0.42\pm0.00$ & $0.42\pm0.00$ & $0.43\pm0.00$ & $0.43\pm0.00$ \\
			\hline
		\end{tabular}
	\end{center}
\end{table}

\subsection{Underwater robot teleoperation with task-parameterized Gaussian mixture model}
\label{sec:DEXROV}

\begin{figure*}
\centering
\includegraphics[width=\textwidth]{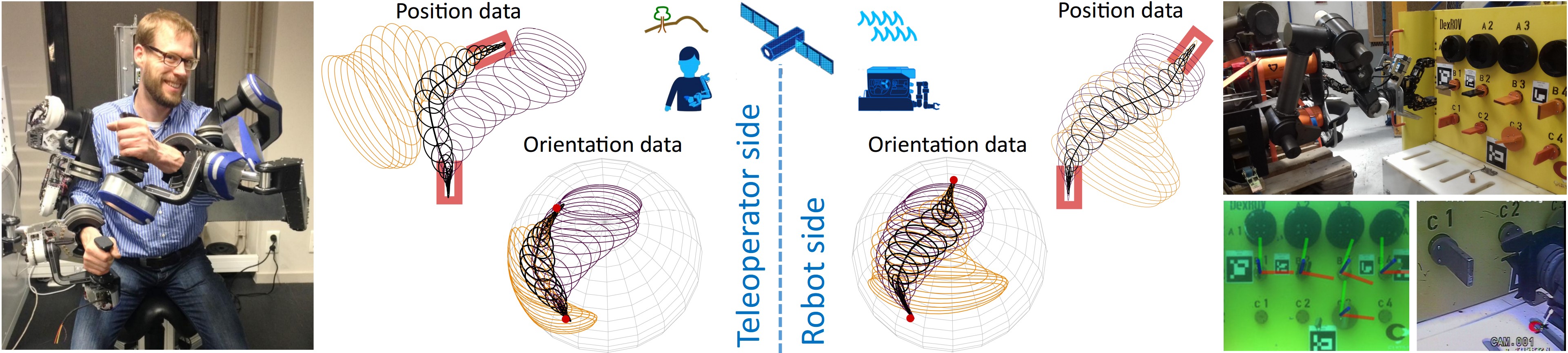}
\caption{
Task-parameterized Gaussian mixture model (TP-GMM) extended to $\mathcal{S}^d$ manifolds within the DexROV project, see main text for details. 
}
\label{fig:TPGMM}
\end{figure*}

The fusion approach presented in Section~\ref{sec:fusion} can be extended to the Gaussian mixture model approach presented in Section~\ref{sec:GMM}. This is particularly useful when mixtures of Gaussians are encoded in different coordinate systems, which need to be fused at reproduction time to satisfy constraints in multiple frames of reference. 

Within the DexROV project~\cite{Birk18RAM}, this task-parameterized Gaussian mixture model (TP-GMM) approach~\cite{Calinon16JIST,Zeestraten17RAL} is used together with the MPC approach presented in Section~\ref{sec:MPC} to teleoperate an underwater robot from distance, with a teleoperator wearing an exoskeleton and visualizing a copy of the robot workspace in a virtual environment. 

Figure~\ref{fig:TPGMM} presents an overview of this application (see also~\cite{Birk18RAM} for a description of this teleoperation approach, and~\cite{Calinon16JIST} for a general description of TP-GMM). Because of the long communication delays between the teleoperator and the robot, the locations of the objects or tools of interest are not the same on the teleoperator side and on the robot side. With a parameterization associated with the locations of objects and tools, we can cope with this discrepancy by adapting locally the movement representation to the position and orientation of the objects/tools, represented as coordinate systems. Figure~\ref{fig:TPGMM} depicts an example with two coordinate systems (with models represented in orange and purple), corresponding respectively to the robot and to a valve that needs to be turned. A motion relative to the valve and to the robot is encoded as Gaussian mixture models (GMM) in the two respective coordinate systems. During teleoperation, each pair of GMMs are rotated and translated according to the current situations on the teleoperator side and on the robot side. Products of Gaussians are then computed at each side to fuse these representations. 

Movement are encoded in this way with both position $\mathbb{R}^3$ and orientation $\mathcal{S}^3$ data (in Fig.~\ref{fig:TPGMM}, a representation with $\mathbb{R}^2$ and $\mathcal{S}^2$ is shown as an illustration). 
For position, the retrieved path in black (and the associated covariances) corresponds to a movement going from the robot to the valve (represented as red U shapes), by taking into account how these different coordinate systems are oriented. We can see that the approaching phase is perpendicular to the coordinate system to properly reach the valve.
For orientation, the retrieved path shows how the orientation of the endeffector change with time. At the beginning of the motion, this orientation relates to the orientation of the robot, while at the end, the orientation of the endeffector matches the orientation of the valve.
In these graphs, the purple and orange ellipsoids depict two GMMs, representing uncertain trajectories with respect to two different frames of reference (red U shapes for position data, and red points on the spheres for orientation data). The black ellipsoids represent the final trajectory and its uncertainty, obtained by fusing the trajectories of the two different frames of reference through products of Gaussians. 
Although the red U shapes and red points are not the same on the teleoperator side and on the robot side, the retrieved paths on the two sides can quickly adapt to these different situations. By using a Riemannian manifold framework, orientations are encoded uniquely in a representation that does not contain singularities. Such an approach is employed in this application on $\mathcal{S}^3$ to learn and retrieve the evolution of robot endeffector orientations, by adapting them to the orientation of objects or tools in the robot workspace.

This approach was successfully tested in field trials in the Mediterranean Sea offshore of Marseille, where 7 extended dives in 4 different sites (8m, 30m, 48m and 100m water depths) were performed with the underwater robot while being connected via satellite to the teleoperation center in Brussels, see~\cite{Birk18RAM} for a general description of the experiment.

\section{FURTHER PERSPECTIVES AND CONCLUSION}
\label{sec:conclusion}

This article showed that a wide range of challenges in robot learning and adaptive control can be recast as statistical modeling and information fusion on Riemannian manifolds. Such an interpretation can avoid potential misuses of algorithms in robotics that might originate from Riemannian geometry but that are treated with a limited view. One such example is to perform all computations in a single tangent space (typically, at the origin of the manifold), instead of considering the closest tangent spaces to avoid distortions. Another example concerns domain adaptation and transfer learning, which require the realignment of data to cope with nonstationarities. For example, sensory data collected by different subjects or throughout several days, which should use the Riemannian notion of \emph{parallel transport} instead of only recentering the data~\cite{Yair18}. 

This article also showed that the combination of statistics and differential geometry offers many research opportunities, and can contribute to recent challenges in robotics. Further work can be organized in two categories. Firstly, the field of robotics is abundant of new techniques proposed by researchers, due to the interdisciplinary aspect and to the richness of problems it involves. The common factor in many of these developments is that they rely on some form of statistics and/or propagation of uncertainty. These models and algorithms are typically developed for standard Euclidean spaces, where an extension to Riemannian manifolds has several benefits to offer.

Secondly, some Riemannian manifolds remain largely underexploited in robotics, despite the fact that some of them are mathematically well understood and characterized by simple closed-form expressions. Grassmann manifolds seem particularly promising to handle problems in robotics with high dimensional datapoints and only few training data, where subspaces are required in the computation to keep the most essential characteristics of the data. It is also promising in problems in which hierarchies are considered (such as inverse kinematics with kinematically redundant robots), because it provides a geometric interpretation of nullspace structures. Other Riemannian manifolds such as hyperbolic manifolds also seem propitious to bring a probabilistic treatment to dynamical systems, tree-based structures, graphs, Toeplitz/Hankel matrices or autoregressive models. Finally, a wide range of metric learning problems in robotics could benefit from a Riemannian geometry treatment.



\bibliographystyle{IEEEtran} 
\bibliography{bib_RAM2020}


\clearpage
\section*{SUPPLEMENTARY MATERIAL}

\subsection*{$\mathcal{S}^d$ manifold}
The exponential and logarithm maps corresponding to the distance
\begin{equation}
	d(\bm{x},\bm{y}) = \arccos(\bm{x}^\trsp \bm{y}), 
	\label{Eq:SphereDist}
\end{equation}
with $\bm{x},\bm{y} \in \mathcal{S}^d$ can be computed as (see also~\cite{Absil07})
\begin{align}
	\bm{y} & = \text{Exp}_{\bm{x}}(\bm{u}) = \bm{x}\cos(\|\bm{u}\|) + \frac{\bm{u}}{\|\bm{u}\|}\sin(\|\bm{u}\|), \\
	\bm{u} & = \text{Log}_{\bm{x}}(\bm{y}) = d(\bm{x},\bm{y}) \, \frac{\bm{y} - \bm{x}^\trsp \bm{y} \, \bm{x}}{\|\bm{y} - \bm{x}^\trsp \bm{y} \, \bm{x}\|}.
	\label{Eq:SphereMaps}
\end{align}
The parallel transport of $\bm{v}\in\mathcal{T}_{\bm{x}}\mathcal{S}^d$ to $\mathcal{T}_{\bm{y}}\mathcal{S}^d$ is given by
\begin{align}
	\Gamma_{\bm{x}\ty{\to}\bm{y}} (\bm{v}) &= \bm{v} -
	\frac{{\text{Log}_{\bm{x}}(\bm{y})}^\trsp \bm{v}}{d(\bm{x},\bm{y})^2} \Big(\text{Log}_{\bm{x}}(\bm{y}) + \text{Log}_{\bm{y}}(\bm{x})\Big).
	\label{Eq:SpherePT}
\end{align}
In some applications, it can be convenient to define the parallel transport with the alternative equivalent form 
\begin{align}
	&\Gamma_{\bm{x}\ty{\to}\bm{y}} (\bm{v}) = \bm{A}_{\bm{x}\ty{\to}\bm{y}} \, \bm{v}, \quad\mathrm{with} \label{Eq:SpherePT2}\\
	&\bm{A}_{\bm{x}\ty{\to}\bm{y}} = -\bm{x}\sin(\|\bm{u}\|)\bm{\overline{u}}^{\trsp} + \bm{\overline{u}}\cos(\|\bm{u}\|)\bm{\overline{u}}^\trsp 
	+ (\bm{I}- \bm{\overline{u}}\,\bm{\overline{u}}^\trsp),\nonumber\\
	&\bm{u} = \text{Log}_{\bm{x}}(\bm{y}), \quad\mathrm{and}\quad \bm{\overline{u}} = \frac{\bm{u}}{\|\bm{u}\|},\nonumber
\end{align}
highlighting the linear structure of the operation.

Corresponding examples in Matlab and C++ can be found in~\cite{pbdlib}, named {\scriptsize\texttt{demo\_Riemannian\_Sd\_*.m}} and {\scriptsize\texttt{demo\_Riemannian\_S2\_*.cpp}}, respectively.

Note that in the above representation, $\bm{u}$ and $\bm{v}$ are described as vectors with $d+1$ elements contained in $\mathcal{T}_{\bm{x}}\mathcal{S}^d$. An alternative representation consists of expressing $\bm{u}$ and $\bm{v}$ as vectors of $d$ elements in the coordinate system attached to $\mathcal{T}_{\bm{x}}\mathcal{S}^d$, see~\cite{Zeestraten17RAL} for details.

\subsection*{$\mathcal{H}^d$ manifold}
The exponential and logarithm maps corresponding to the distance
\begin{equation}
	d(\bm{x},\bm{y}) = \text{arccosh}\Big(\!-\!\langle\bm{x}, \bm{y}\rangle_\text{M}\!\Big),
	\label{Eq:HyperbolicDist}
\end{equation}
with $\bm{x},\bm{y} \in \mathcal{H}^d$ can be computed as
\begin{align}
	\bm{y} & = \text{Exp}_{\bm{x}}(\bm{u}) = \bm{x}\cosh(\|\bm{u}\|_\text{M}) + \frac{\bm{u}}{\|\bm{u}\|_\text{M}}\sinh(\|\bm{u}\|_\text{M}), \\ 
	\bm{u} & = 
	\text{Log}_{\bm{x}}(\bm{y}) = d(\bm{x},\bm{y}) \, \frac{\bm{y} + \langle \bm{x}, \bm{y}\rangle_\text{M} \, \bm{x}}{{\|\bm{y} + \langle \bm{x}, \bm{y}\rangle_\text{M} \, \bm{x}\|}_\text{M}},
	\label{Eq:HyperbolicMaps}
\end{align}
by using the Minkowski inner product $\langle \bm{x},\bm{y}\rangle_\text{M}=\bm{x}^\trsp \left(
\begin{smallmatrix}
\bm{I} &\bm{0} \\
\bm{0} & -1 \\
\end{smallmatrix}
\right)\bm{y}$ and norm $\!\|\bm{x}\|_\text{M} = \sqrt{\langle \bm{x},\bm{x}\rangle_\text{M}}$.
The parallel transport of $\bm{v}\in\mathcal{T}_{\bm{x}}\mathcal{H}^d$ to $\mathcal{T}_{\bm{y}}\mathcal{H}^d$ is given by
\begin{align}
	\Gamma_{\bm{x}\ty{\to}\bm{y}} (\bm{v}) &= \bm{v} -
	\frac{\langle \text{Log}_{\bm{x}}(\bm{y}) , \bm{v}\rangle_\text{M}}{d(\bm{x},\bm{y})^2} \Big(\text{Log}_{\bm{x}}(\bm{y}) + \text{Log}_{\bm{y}}(\bm{x})\Big).
	\label{Eq:HyperbolicPT}
\end{align}

Corresponding examples in Matlab can be found in~\cite{pbdlib}, named {\scriptsize\texttt{demo\_Riemannian\_Hd\_*.m}}.

\subsection*{$\mathcal{S}^d_{++}$ manifold}
For an affine-invariant distance between $\bm{X},\bm{Y}\in\mathcal{S}_{++}^d$
\begin{equation}
	d(\bm{X},\bm{Y}) = {\big\|\log(\bm{X}^{-\frac{1}{2}}\bm{Y}\bm{X}^{-\frac{1}{2}})\big\|}_\text{F}, 
	\label{Eq:SPDdist}
\end{equation}
the exponential and logarithmic maps on the SPD manifold can be computed as (see also~\cite{Pennec06})
\begin{align}
	\bm{Y} & = \text{Exp}_{\bm{X}}(\bm{U}) = \bm{X}^{\frac{1}{2}}\exp\big(\bm{X}^{-\frac{1}{2}}\bm{U}\bm{X}^{-\frac{1}{2}}\big)\bm{X}^{\frac{1}{2}}, \\
	\bm{U} & = \text{Log}_{\bm{X}}(\bm{Y}) = \bm{X}^{\frac{1}{2}}\log\big(\bm{X}^{-\frac{1}{2}}\bm{Y}\bm{X}^{-\frac{1}{2}}\big)\bm{X}^{\frac{1}{2}}.
	\label{Eq:SPDmaps}
\end{align}
The parallel transport of $\bm{V}\in\mathcal{T}_{\bm{X}}\mathcal{S}_{++}^d$ to $\mathcal{T}_{\bm{Y}}\mathcal{S}_{++}^d$ is given by
\begin{equation}
	\Gamma_{\bm{X}\ty{\to}\bm{Y}} (\bm{V}) = \bm{A}_{\bm{X}\ty{\to}\bm{Y}} \, \bm{V} \, \bm{A}_{\bm{X}\ty{\to}\bm{Y}}^\trsp,
	\,\mathrm{with}\,
	\bm{A}_{\bm{X}\ty{\to}\bm{Y}} = \bm{Y}^{\frac{1}{2}}\bm{X}^{-\frac{1}{2}}.
	\label{eq:spdPT}
\end{equation}

Corresponding examples in Matlab and C++ can be found in~\cite{pbdlib}, named {\scriptsize\texttt{demo\_Riemannian\_SPD\_*.m}} and {\scriptsize\texttt{demo\_Riemannian\_SPD\_*.cpp}}, respectively.

\subsection*{$\mathcal{G}^{d,p}$ manifold}
For the arc length distance between two points $\bm{X},\bm{Y}\in\mathcal{G}^{d,p}$ 
\begin{equation}
	d(\bm{X},\bm{Y}) = {\|\arccos(\bm{\sigma})\|}_2, 
	\quad\mathrm{with}\quad
	\bm{\sigma} = \mathrm{vec}(\bm{\Sigma}), 
	\label{Eq:GrassmannDist}
\end{equation}
computed with the singular value decomposition (SVD) $\bm{X}^\trsp\bm{Y} = \bm{U}\bm{\Sigma}\bm{V}^\trsp$, the exponential and logarithm map of the Grassmann manifold are given by~\cite{Edelman98} 
\begin{equation}
	\bm{Y} = \text{Exp}_{\bm{X}}(\bm{H}) = (\begin{matrix}\bm{X}\bm{V}& \bm{U}\end{matrix})
	\left(\begin{matrix}
	\cos\bm{\Sigma} \\ \sin{\bm{\Sigma}}
	\end{matrix} \right) \bm{V}^\trsp, 
	\label{Eq:GrassmanMaps}
\end{equation}
computed with the SVD $\bm{H} = \bm{U}\bm{\Sigma}\bm{V}^\trsp$, where
\begin{equation}
	\bm{H} = \text{Log}_{\bm{X}}(\bm{Y}) = \bm{U}\arctan(\bm{\Sigma})\bm{V}^\trsp
\end{equation}
is computed with the SVD $(\bm{I}-\bm{X}\bm{X}^\trsp) \bm{Y} {(\bm{X}^\trsp\bm{Y})}^{-1} = \bm{U}\bm{\Sigma}\bm{V}^\trsp$.
The parallel transport of $\bm{G}\in\mathcal{T}_{\bm{X}}\mathcal{G}^{d,p}$ to $\mathcal{T}_{\bm{Y}}\mathcal{G}^{d,p}$ corresponds to
\begin{align}
	\Gamma_{\bm{X}\ty{\to}\bm{Y}} (\bm{G}) &=  \Big( (\begin{matrix}\bm{X}\bm{V}& \bm{U}\end{matrix})
	\left(\begin{matrix}
	-\sin\bm{\Sigma} \\ \cos{\bm{\Sigma}}
	\end{matrix} \right)\bm{U}^\trsp + (\bm{I}-\bm{U}\bm{U}^\trsp) \Big) \bm{G},
	\label{Eq:GrassmanPT}
\end{align}
computed with the SVD $\text{Log}_{\bm{X}}(\bm{Y}) = \bm{U}\bm{\Sigma}\bm{V}^\trsp$.

Corresponding examples in Matlab can be found in~\cite{pbdlib}, named {\scriptsize\texttt{demo\_Riemannian\_Gdp\_*.m}}.

\subsection*{Computation of transfer matrices $\bm{S}_{\bm{u}}$ and $\bm{S}_{\bm{x}}$ in MPC}
The MPC problem of estimating velocity commands $\bm{u}_t\!\in\!\mathbb{R}^{d}$ with a discrete linear dynamical system $\bm{x}_{t+1} = f(\bm{x}_t,\bm{u}_t)$ can be solved by linearization with
\begin{equation}
	\bm{x}_{t+1} = \bm{A}_t(\bm{x}_t,\bm{u}_t) \; \bm{x}_t + \bm{B}_t(\bm{x}_t,\bm{u}_t) \; \bm{u}_t,
	\label{eq:AB}
\end{equation}
and expressing all future states $\bm{x}_t$ as an explicit function of the state $\bm{x}_1$. By writing
\begin{align*}
	\bm{x}_{2} &= \bm{A}_1 \bm{x}_1 + \bm{B}_1 \bm{u}_1 ,\\
	\bm{x}_{3} &= \bm{A}_2 \bm{x}_2 + \bm{B}_2 \bm{u}_2 = \bm{A}_2 (\bm{A}_1 \bm{x}_1 + \bm{B}_1 \bm{u}_1) + \bm{B}_2 \bm{u}_2 ,\\[-2mm]
	&\vdots\\[-2mm]
	\bm{x}_{T} &= \prod_{t=1}^{T-1} \bm{A}_{T-t} \bm{x}_1 \;+ \prod_{t=1}^{T-2} \bm{A}_{T-t} \bm{B}_1 \bm{u}_1 \;+ \\
	& \hspace{22mm} \prod_{t=1}^{T-3} \bm{A}_{T-t} \bm{B}_2 \bm{u}_2 \;+\; \cdots \;+\; \bm{B}_{T-1} \bm{u}_{T-1},
\end{align*}
in a matrix form, we get an expression of the form $\bm{x}=\bm{S}_{\bm{x}}\bm{x}_1+\bm{S}_{\bm{u}}\bm{u}$, with
\begin{equation*}
	\underbrace{
 	\begin{bmatrix}
 	\bm{x}_1\\
 	\bm{x}_2\\
 	\bm{x}_3\\
 	\vdots\\
 	\bm{x}_T
 	\end{bmatrix}}_{\bm{x}}
 	=
 	\underbrace{
 	\begin{bmatrix}
 	\bm{I}\\
 	\bm{A}_1\\
 	\bm{A}_2\bm{A}_1\\
 	\vdots\\
 	\prod_{t=1}^{T-1} \bm{A}_{T-t}
 	\end{bmatrix}}_{\bm{S}_{\bm{x}}}
 	\bm{x}_1
	\quad + \hspace{38mm}
\end{equation*}
\begin{equation*}
 	\underbrace{
 	\begin{bmatrix}
 	\bm{0} & \bm{0} & \cdots & \bm{0} \\
 	\bm{B}_1 & \bm{0} & \cdots & \bm{0} \\
	\bm{A}_2\bm{B}_1 & \bm{B}_2 & \cdots & \bm{0} \\
	\vdots & \vdots & \ddots & \vdots \\
	\prod_{t=1}^{T-2} \bm{A}_{T-t} \bm{B}_1 & \prod_{t=1}^{T-3} \bm{A}_{T-t} \bm{B}_2 & \cdots & \bm{B}_{T-1}
	\end{bmatrix}}_{\bm{S}_{\bm{u}}}
	\underbrace{
	\begin{bmatrix}
	\bm{u}_1\\
 	\bm{u}_2\\
 	\vdots\\
 	\bm{u}_{T\!-\!1}
 	\end{bmatrix}}_{\bm{u}},
\end{equation*}
where $\bm{S}_{\bm{x}}\!\in\!\mathbb{R}^{dT\times d}$, $\bm{x}_1\!\in\!\mathbb{R}^d$, $\bm{S}_{\bm{u}}\!\in\!\mathbb{R}^{dT\times d(T-1)}$ and $\bm{u}\!\in\!\mathbb{R}^{d(T-1)}$. 

Corresponding examples in Matlab and C++ can be found in~\cite{pbdlib}, named {\scriptsize\texttt{demo\_MPC\_*.m}} and {\scriptsize\texttt{demo\_MPC\_*.cpp}}, respectively.

\end{document}